\documentclass{article}
\usepackage{spconf,amsmath,graphicx,hyperref}
\usepackage{booktabs}
\usepackage{multirow}
\usepackage{url}
\title{LBSNet: Lightweight Brain State Modeling via Learnable Spectral Prototypes for EEG-Based Fatigue Detection}
\name{Kun Jia, Zhenyu Cao, Hanlin Yu, Zeyuan Li, Yixuan Liu, Zhongye Xia, Sen Song$^{\star}$, Zhiyuan Ma$^{\star}$%
\thanks{$^{\star}$Corresponding authors: Zhiyuan Ma (\protect\nolinkurl{ma-zy25@mails.tsinghua.edu.cn}) and Sen Song (\protect\nolinkurl{songsen@tsinghua.edu.cn}).
This work was supported in part by the National Natural Science Foundation of China under Grants 2025ZD0215701, T2341003, and U2336214, in part by the Beijing Natural Science Foundation under Grant L257019, and in part by the Tsinghua IDG/McGovern "Brain+X" Seed Grant Doctoral and Postdoctoral Program.}}

\address{School of Biomedical Engineering, Tsinghua Medicine, Tsinghua University, Beijing, China}
\usepackage{graphicx}
\begin{document}
\ninept
\maketitle
\begin{abstract}
EEG-based fatigue detection requires modeling spectral information across bands and its temporal variation. Recent methods combine multi-band modeling with adaptive temporal processing, yet how spectral structure across bands should influence the contribution of each temporal interval remains insufficiently explored. We propose LBSNet, a lightweight network that uses prototype-based brain-state modeling to link local spectral structure with temporal aggregation. LBSNet extracts compact multi-band features through band-specific temporal convolution, shared spatial projections, and cross-band recalibration. Within each window, relative spectral activity is encoded as a normalized descriptor and softly matched to learnable spectral prototypes. Prototype assignments and learned prototype importance jointly determine temporal weights for power aggregation, allowing the spectral pattern of each window to guide its contribution to the prediction. Subject-independent experiments on SEED-VIG and SADT show that LBSNet outperforms seven baselines while maintaining low computational cost. Ablation and visualization analyses support prototype-guided temporal weighting, while the learned prototypes reveal a gradual redistribution of relative cross-band spectral states with increasing fatigue. The code is publicly available at https://github.com/zy0929/LBSNet.
\end{abstract}

\section{Introduction}
\label{sec:introduction}


Driver fatigue impairs vigilance, sustained attention, and responsiveness, increasing the risk of driving errors and accidents \cite{mollicone2019predicting, gibbings2022eeg}. Electroencephalography (EEG) provides a noninvasive measure of brain activity and has been widely used for objective fatigue assessment \cite{borghini2014measuring, hussein2023driver}. Prior studies have consistently linked fatigue and vigilance to changes in spectral activity across canonical EEG bands, and these spectral characteristics vary over the course of driving \cite{tran2020influence, SEED-VIG}. EEG-based fatigue detection therefore requires modeling both fatigue-related spectral activity and its temporal variation.

Early EEG-based fatigue studies combined spectral features with temporal modeling, with Zheng and Lu \cite{SEED-VIG} incorporating temporal dependencies into vigilance estimation using frequency-specific features. Subsequent convolutional approaches moved toward end-to-end learning. InterpretableCNN \cite{InterpretableCNN} learns fatigue-related features for cross-subject drowsiness recognition and reveals alpha spindles through interpretation analysis. Attention-based methods further strengthened spectral-temporal modeling. SACC-CapsNet \cite{SACC-CapsNet} emphasizes temporal components and channels, while TFormer \cite{TFormer} integrates temporal and frequency-domain patterns through cross-attention. Recent models combine frequency-specific modeling with adaptive temporal processing. Multi-band modeling and temporal weighting capture different aspects of fatigue-related EEG, but existing designs do not explicitly link the local spectral structure of an interval to its temporal contribution \cite{FAST-Net, MB-STFormer}. Overall, EEG-based fatigue detection has progressed toward representation learning and adaptive modeling across frequency and time.

Multi-band modeling and temporal weighting capture different aspects of fatigue-related EEG, but they do not define how the spectral structure of an interval should influence its temporal contribution \cite{FAST-Net, MB-STFormer}. This issue is important because fatigue has been characterized not only by changes within individual frequency bands, but also by relative band power and relationships among bands \cite{li2022electroencephalogram}. Accordingly, fatigue-related information within a local interval may be reflected in joint patterns of activity across frequency bands \cite{arif2023driving}. This raises the question of whether temporal aggregation can be guided by such local spectral patterns. Grouping similar local spectral patterns into shared states and learning their relevance to fatigue would provide a link between spectral structure and temporal weighting.

To this end, we propose LBSNet, a lightweight network for EEG-based fatigue detection. Its central idea is prototype-based brain-state modeling, which links local patterns of relative activity across frequency bands to temporal weighting. LBSNet first decomposes EEG into five canonical frequency bands, applies band-specific temporal convolution, and uses spatial projections shared across bands to obtain compact features. Cross-band recalibration then adjusts the relative contribution of the five bands based on global spectral activity. Within each temporal window, envelope-based band activity is transformed into a normalized relative spectral descriptor and softly matched to the learned prototypes. The resulting prototype assignments are combined with learned prototype importance to derive temporal weights, which are used to aggregate component-wise power into a compact representation for classification. The main contributions are summarized as follows:
\begin{enumerate}
    \item We introduce prototype-based brain-state modeling that represents local relative spectral patterns with shared learnable prototypes and combines prototype assignments with learned importance to guide temporal aggregation.
    \item We develop a lightweight architecture with shared spatial projections, low-dimensional state matching, and a compact classifier, achieving a favorable accuracy-efficiency trade-off with a small parameter budget and low computational cost.
    \item We validate LBSNet on SEED-VIG and SADT and show that the learned spectral states exhibit a gradual redistribution with increasing fatigue, providing empirical support for prototype-guided temporal weighting.
\end{enumerate}

\begin{figure*}[t]
    \centering
    \includegraphics[width=\textwidth, trim={15 15 15 15},clip]{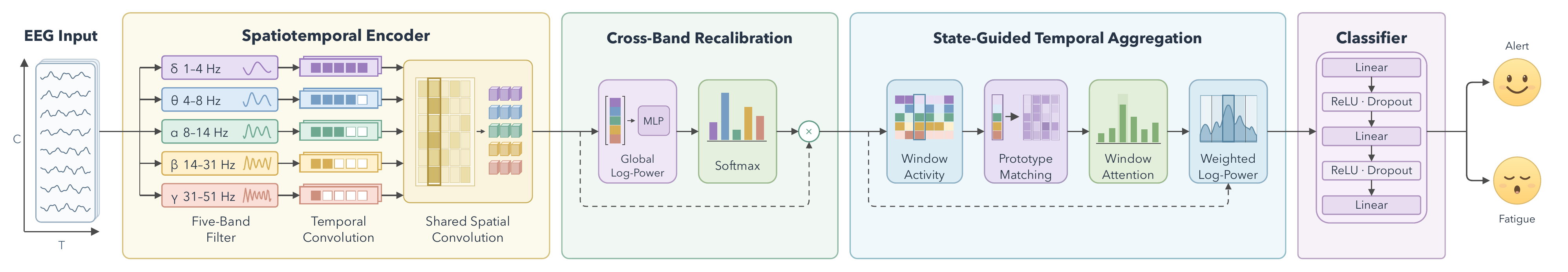}
    \caption{Overview of LBSNet, where learnable spectral prototypes represent local relative spectral patterns and guide temporal aggregation.}
    \label{fig:lbsnet}
\end{figure*}

\section{Method}
\label{sec:method}

As illustrated in Fig.~\ref{fig:lbsnet}, LBSNet comprises
a spatiotemporal encoder, cross-band recalibration, and
state-guided temporal aggregation. Starting from multichannel
EEG signals, LBSNet extracts spatiotemporal features from
five frequency bands, recalibrates their relative contributions,
and uses learnable spectral prototypes to guide temporal
aggregation, enabling lightweight modeling of latent brain states
for EEG-based fatigue classification.

\subsection{Spatiotemporal Encoder}
\label{sec:encoder}

The spatiotemporal encoder extracts five-band EEG features
through temporal and shared spatial convolutions.
For an input $X\in\mathbf{R}^{C\times T}$, with $C$ electrodes
and $T$ samples, a fixed filter bank decomposes the signal
into five frequency bands:
\begin{equation}
    X^{(b)}=\mathcal{B}_b(X),
    \qquad b=1,\ldots,5,
    \label{eq:encoder-filterbank}
\end{equation}
where $\mathcal{B}_b$ applies Butterworth filtering independently
to each electrode, yielding the band signal $X^{(b)}$.
The passbands are 1--4, 4--8, 8--14, 14--31, and 31--51~Hz.

Each band undergoes temporal convolution:
\begin{equation}
    U_{b,c,:}
    =
    \operatorname{Conv1D}\!\left(
        X^{(b)}_{c,:};\boldsymbol{\theta}_b
    \right),
    \label{eq:encoder-temporal}
\end{equation}
where $c$ indexes electrodes, the colon denotes the temporal
sequence, and $\boldsymbol{\theta}_b$ is a band-specific kernel
shared across electrodes. These bias-free convolutions use kernel lengths
$(125,63,31,15,7)$ at 128~Hz and proportionally adjusted
odd lengths at other sampling rates.
Symmetric zero padding preserves the temporal length,
yielding the temporally filtered signals
$U\in\mathbf{R}^{5\times C\times T}$.

Before spatial convolution, the signals are centered across
electrodes and the spatial convolution kernels are normalized:
\begin{equation}
    \begin{aligned}
        V_{bct}
        &= U_{bct}-\frac{1}{C}\sum_{j=1}^{C}U_{bjt},\\
        \widehat{\mathbf w}_m
        &= \mathcal{N}(\mathbf w_m),
        \qquad m=1,\ldots,M,
    \end{aligned}
    \label{eq:encoder-normalization}
\end{equation}
where $V$ contains the centered signals, $j$ indexes electrodes,
and $t$ indexes samples. The vector
$\mathbf w_m\in\mathbf{R}^{C}$ is the $m$th spatial convolution
kernel, and $M$ is the number of output features per band.
The operator $\mathcal{N}$ removes the kernel mean and applies
$\ell_2$ normalization.

Shared spatial convolution then produces
\begin{equation}
    F_{bmt}=\sum_{c=1}^{C}\widehat w_{mc}V_{bct},
    \label{eq:encoder-spatial}
\end{equation}
where $\widehat w_{mc}$ is the coefficient for electrode $c$
in normalized kernel $m$, and $F_{bmt}$ is its response in
band $b$ at sample $t$. The same kernels span all electrodes
and are shared across bands. The output
$F\in\mathbf{R}^{5\times M\times T}$ retains the band dimension
and is passed to cross-band recalibration for joint
band-weight estimation.

\subsection{Cross-Band Recalibration}
\label{sec:band-recalibration}

For the band-specific encoder output $F$, cross-band
recalibration first computes global log-power descriptors
for the five frequency bands:
\begin{equation}
    d_b
    =
    \log\!\left(
        \frac{1}{MT}
        \sum_{m=1}^{M}\sum_{t=1}^{T}F_{bmt}^{2}
    \right),
    \label{eq:band-descriptor}
\end{equation}
where $d_b$ is the global log-power of band $b$, obtained
by averaging power over spatial features and temporal
samples and then taking the logarithm.
The descriptors form $\mathbf d=[d_1,\ldots,d_5]^\top$.

Band weights are jointly estimated as
\begin{equation}
    \boldsymbol{\alpha}
    =
    \operatorname{softmax}\!\left(g(\mathbf d)\right),
    \label{eq:band-weights}
\end{equation}
where $g$ is a two-layer MLP with an intermediate ReLU,
and $\boldsymbol{\alpha}$ contains five positive weights
summing to one. Each weight is conditioned on the
descriptors of all five bands.

The encoder features are then recalibrated:
\begin{equation}
    G_{bmt}=\sqrt{\alpha_b}\,F_{bmt},
    \label{eq:band-recalibration}
\end{equation}
where $\alpha_b$ is the weight for band $b$, shared across
its spatial features and temporal samples.
Square-root scaling weights the corresponding power
by $\alpha_b$. The recalibrated output
$G\in\mathbf{R}^{5\times M\times T}$ is passed to
state-guided temporal aggregation.

\subsection{State-Guided Temporal Aggregation}
\label{sec:state-aggregation}

Following cross-band recalibration, state-guided temporal
aggregation weights temporal windows according to their
latent spectral patterns. The recalibrated features $G$
are divided into $N=T/L$ non-overlapping windows of $L$
samples. Band activity within each window is computed
from the analytic signal as
\begin{equation}
    a_{kb}
    =
    \frac{1}{2ML}
    \sum_{m=1}^{M}\sum_{t\in\mathcal I_k}
    \left[
        G_{bmt}^{2}
        + \mathcal{H}(G_{bm,:})_t^{2}
    \right],
    \label{eq:window-activity}
\end{equation}
where $\mathcal I_k$ contains the sample indices of window
$k$, $\mathcal{H}$ denotes the Hilbert transform along the
full temporal sequence, and $a_{kb}$ measures the activity
of band $b$ in that window.

The window activity vector is transformed into a normalized
spectral descriptor that captures relative activity across bands:
\begin{equation}
    \mathbf z_k
    =
    \mathcal{N}\!\left(\log\mathbf a_k\right),
    \qquad
    \mathbf a_k=[a_{k1},\ldots,a_{k5}]^\top,
    \label{eq:window-descriptor}
\end{equation}
where $\mathbf z_k\in\mathbf{R}^{5}$, the logarithm is
applied elementwise, and $\mathcal{N}$ performs mean removal
and $\ell_2$ normalization across bands.
This emphasizes relative spectral patterns and reduces
sensitivity to variations in the overall magnitude of the EEG signal.

Each window descriptor is matched against a set of $R$
learnable spectral prototypes to obtain soft state assignments:
\begin{equation}
    \begin{aligned}
        \widehat{\mathbf p}_r
        &= \mathcal{N}(\mathbf p_r),\\
        q_{rk}
        &=
        \frac{
            \exp\!\left(
                \widehat{\mathbf p}_r^\top\mathbf z_k/\tau_p
            \right)
        }{
            \sum_{s=1}^{R}
            \exp\!\left(
                \widehat{\mathbf p}_s^\top\mathbf z_k/\tau_p
            \right)
        },
    \end{aligned}
    \label{eq:prototype-matching}
\end{equation}
where $\mathbf p_r\in\mathbf{R}^{5}$ is prototype $r$,
$\widehat{\mathbf p}_r$ is its normalized form, $q_{rk}$
is the soft assignment of window $k$ to that prototype,
and $\tau_p$ controls assignment sharpness.
The prototypes represent latent spectral patterns learned
jointly with the network.

The assignments are combined with learnable prototype
importance to obtain temporal weights:
\begin{equation}
    \beta_k
    =
    \frac{
        \exp\!\left(
            \sum_{r=1}^{R}\eta_r q_{rk}/\tau_t
        \right)
    }{
        \sum_{j=1}^{N}
        \exp\!\left(
            \sum_{r=1}^{R}\eta_r q_{rj}/\tau_t
        \right)
    },
    \label{eq:window-attention}
\end{equation}
where $\eta_r$ denotes the importance of prototype $r$,
$\tau_t$ controls temporal weighting sharpness, and
$\beta_k$ is the weight of window $k$, with
$\sum_k\beta_k=1$. Thus, each window contributes according
to both its state assignments and the learned importance
of those states.

The temporal weights are then used to aggregate squared
recalibrated features into a weighted log-power representation:
\begin{equation}
    h_{bm}
    =
    \log\!\left[
        \frac{1}{M}
        \sum_{k=1}^{N}\beta_k
        \left(
            \frac{1}{L}
            \sum_{t\in\mathcal I_k}G_{bmt}^{2}
        \right)
    \right],
    \label{eq:weighted-log-power}
\end{equation}
where $h_{bm}$ is the weighted log-power of spatial feature
$m$ in band $b$. The same temporal weights are shared
across bands and spatial features. The
matrix $\mathbf H=[h_{bm}]\in\mathbf{R}^{5\times M}$
is flattened into a $5M$-dimensional representation
and passed to a three-layer classifier with hidden
dimensions 32 and 16. ReLU and dropout follow each
hidden layer, and the final linear layer produces
the class logits.

\section{Experiments and Results}
\subsection{Datasets and Baselines}

We evaluate our method on two EEG benchmarks for driver
fatigue detection: \textbf{SEED-VIG}~\cite{SEED-VIG} and
\textbf{SADT}~\cite{SADT2022}. \textbf{SEED-VIG} contains
EEG recordings from 23 subjects during simulated driving.
We retain 17 channels, downsample the signals from 1,000
to 200 Hz, and segment them into 20,355 non-overlapping
8-s samples. Vigilance is annotated using PERCLOS derived
from eye-tracking recordings, with 0.35 as the threshold
between alert and fatigued states. \textbf{SADT} is a balanced subset \cite{SADT2022} of the sustained-attention driving dataset \cite{SADT}, comprising 2,022 3-s samples from 11 subjects at 128 Hz. We use all 30 EEG channels. The provided alert and drowsy labels were derived from local and global reaction-time criteria, considering both the current reaction time and recent response history.

For comparison, we select seven representative baselines.
The convolutional baselines include EEGNet~\cite{EEGNet}
and TCNet~\cite{TCNet}. The attention-based baselines
include Conformer~\cite{Conformer}, Deformer~\cite{Deformer},
BiConformer~\cite{BiConformer}, DSAINet~\cite{DSAINet},
and MB-STFormer~\cite{MB-STFormer}.

\subsection{Experimental Setup}

Experiments are conducted on a server equipped with eight
NVIDIA GeForce RTX 4090 GPUs. We use subject-wise five-fold
cross-validation, where all samples from each subject are
assigned to a single fold; three folds are used for training,
one for validation, and one for testing. The model with
the highest validation accuracy is selected for final
evaluation. We set the temporal window length of LBSNet
to 0.5\,s for both datasets, with $L$ denoting the
corresponding number of samples. The number of spectral
prototypes is set to $R=4$ for SADT and $R=2$ for
SEED-VIG, while the number of spatial features per band
is set to $M=2$ and $M=4$, respectively. All experiments are repeated with five random seeds,
and the results are reported as mean $\pm$ standard deviation
across seeds.

\subsection{Performance Comparison}

Table~\ref{tab:comparison} compares the performance of LBSNet with seven baselines on SEED-VIG and SADT. LBSNet achieves the highest accuracy and F1-score on both datasets, with larger gains observed on SADT. Compared with the strongest baseline MB-STFormer, LBSNet improves accuracy/F1-score by 1.26/1.21 percentage points on SEED-VIG and 2.68/2.77 percentage points on SADT.

\begin{table}[t]
\centering
\caption{Performance comparison (\%) on SEED-VIG and SADT.}
\label{tab:comparison}
\setlength{\tabcolsep}{2.3pt}
\renewcommand{\arraystretch}{1.0}
\begin{tabular}{@{}lcccc@{}}
\toprule
\multirow[c]{2}{*}{\raisebox{-0.5ex}{Method}}
& \multicolumn{2}{c}{SEED-VIG}
& \multicolumn{2}{c}{SADT} \\
\cmidrule(lr){2-3}
\cmidrule(lr){4-5}
& Accuracy & F1-score
& Accuracy & F1-score \\
\midrule
EEGNet
& 56.92$\pm$1.63 & 53.71$\pm$1.16
& 67.42$\pm$2.76 & 66.62$\pm$3.10 \\

TCNet
& 57.01$\pm$2.72 & 54.26$\pm$2.52
& 68.75$\pm$1.66 & 68.11$\pm$1.89 \\

Conformer
& 58.08$\pm$2.03 & 55.09$\pm$2.47
& 70.38$\pm$1.66 & 69.72$\pm$2.02 \\

Deformer
& 58.53$\pm$2.72 & 55.21$\pm$2.40
& 70.61$\pm$2.43 & 69.91$\pm$3.07 \\

BiConformer
& 58.70$\pm$1.76 & 55.92$\pm$1.58
& 70.71$\pm$1.74 & 70.18$\pm$1.88 \\

DSAINet
& 59.36$\pm$1.96 & 58.19$\pm$2.41
& 70.96$\pm$2.47 & 70.46$\pm$2.60 \\

MB-STFormer
& 61.47$\pm$2.02 & 60.30$\pm$2.21
& 71.06$\pm$2.45 & 70.53$\pm$2.43 \\

\textbf{LBSNet}
& \textbf{62.73$\pm$1.82} & \textbf{61.51$\pm$1.39}
& \textbf{73.74$\pm$1.38} & \textbf{73.30$\pm$1.43} \\
\bottomrule
\end{tabular}
\end{table}

\begin{table}[t]
\centering
\caption{Ablation results (\%) on SEED-VIG and SADT.}
\label{tab:ablation}
\setlength{\tabcolsep}{1.3pt}
\renewcommand{\arraystretch}{1.0}
\begin{tabular}{@{}lcccc@{}}
\toprule
\multirow[c]{2}{*}{\raisebox{-0.5ex}{Method}}
& \multicolumn{2}{c}{SEED-VIG}
& \multicolumn{2}{c}{SADT} \\
\cmidrule(lr){2-3}
\cmidrule(lr){4-5}
& Accuracy & F1-score
& Accuracy & F1-score \\
\midrule
\begin{tabular}[c]{@{}l@{}}
Uniform Pooling
\end{tabular}
& 60.25$\pm$1.49 & 58.61$\pm$2.66
& 71.96$\pm$2.24 & 72.49$\pm$2.48 \\

\begin{tabular}[c]{@{}l@{}}
MLP Attention
\end{tabular}
& 60.63$\pm$2.37 & 58.01$\pm$2.46
& 72.55$\pm$0.96 & 71.99$\pm$1.00 \\

\begin{tabular}[c]{@{}l@{}}
w/o CBR
\end{tabular}
& 61.49$\pm$1.42 & 58.07$\pm$2.23
& 72.23$\pm$1.37 & 72.68$\pm$1.54 \\

\textbf{LBSNet}
& \textbf{62.73$\pm$1.82} & \textbf{61.51$\pm$1.39}
& \textbf{73.74$\pm$1.38} & \textbf{73.30$\pm$1.43} \\
\bottomrule
\end{tabular}
\end{table}

\subsection{Ablation Analysis}
\label{sec:ablation}

To assess the components of LBSNet, we examine the role of state-guided temporal aggregation, whether prototype matching offers benefits beyond learnable temporal weighting alone, and the contribution of cross-band recalibration (CBR). As shown in Table~\ref{tab:ablation}, replacing the learned temporal weights with Uniform Pooling reduces accuracy/F1-score by 2.48/2.90 points on SEED-VIG and 1.78/0.81 points on SADT, showing the benefit of state-guided temporal aggregation. Replacing prototype matching with parameter-matched MLP Attention also reduces performance. Compared with this variant, LBSNet improves accuracy/F1-score by 2.10/3.50 points on SEED-VIG and 1.19/1.31 points on SADT. MLP Attention itself yields only small accuracy gains over Uniform Pooling and lower F1-scores on both datasets, indicating that learnable temporal weighting alone does not explain the improvement. Finally, removing CBR reduces both accuracy and F1-score on both datasets, showing that cross-band recalibration also contributes to the performance.

\begin{figure}[t]
    \centering
    \includegraphics[width=1\linewidth, trim={5 5 0 0},clip]{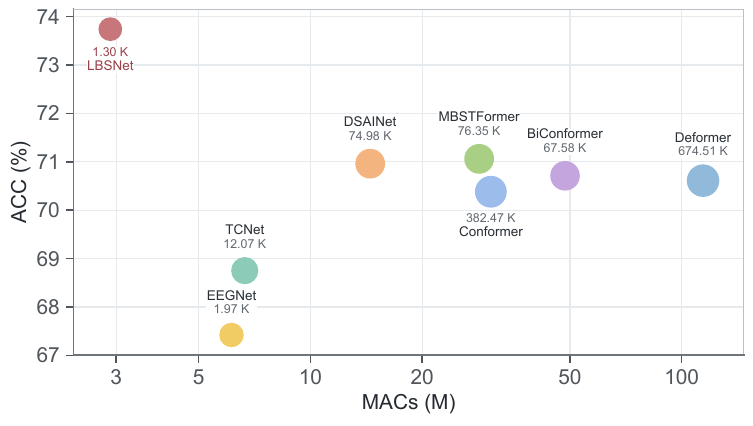}
    \caption{Comparison of model accuracy and computational efficiency on SADT. Each bubble corresponds to a model. The x-axis shows MACs (M) on a logarithmic scale, while the y-axis reports ACC (\%). Bubble area reflects the logarithm of the number of trainable parameters, and the labels give the corresponding parameter counts.}
    \label{fig:efficiency}
\end{figure}

\begin{figure}[t]
    \centering
    \includegraphics[width=1\linewidth, trim={0 0 0 0},clip]{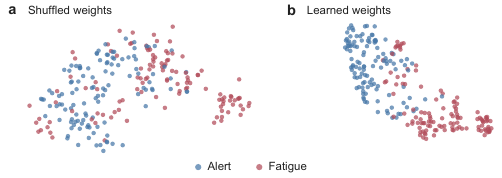}
    \caption{t-SNE visualization of representations produced by temporal aggregation on SADT. (a) Shuffled weights. (b) Learned weights.}
    \label{fig:tsne}
\end{figure}

\begin{figure}[t]
    \centering
    \includegraphics[width=1\linewidth, trim={0 5 0 5},clip]{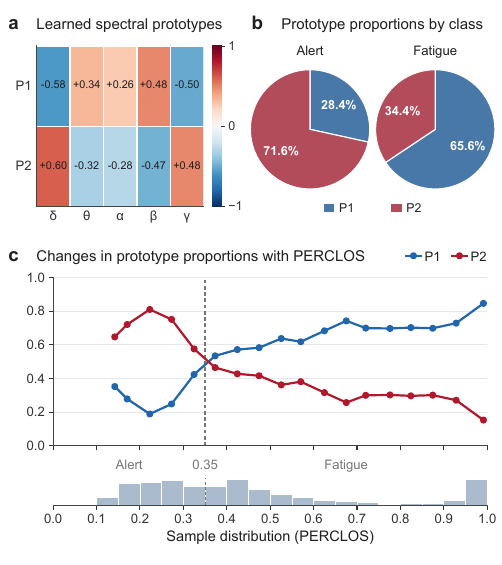}
    \caption{Analysis of learned spectral prototypes on SEED-VIG. (a) Relative spectral patterns of the learned prototypes across the five frequency bands. (b) Prototype proportions for alert and fatigue samples. (c) Changes in prototype proportions across PERCLOS values. The dashed line marks the 0.35 threshold between alert and fatigue, and the histogram shows the sample distribution.}
    \label{fig:prototype}
\end{figure}

\subsection{Computational Efficiency Analysis}
\label{sec:computational_efficiency_analysis}

We compare models on SADT in terms of accuracy, multiply-accumulate operations (MACs), and trainable parameters. Fig.~\ref{fig:efficiency} shows that LBSNet achieves the highest mean accuracy of 73.74\% while requiring only 2.89 M MACs and 1.30 K trainable parameters, both the lowest among the compared models. Compared with EEGNet, the lightest baseline, LBSNet reduces MACs by 52.9\% and the parameter count by 34.0\%, while achieving higher accuracy. Compared with MB-STFormer, the strongest baseline, LBSNet improves accuracy by 2.68 percentage points while reducing MACs by 89.9\% and the parameter count by 98.3\%. The efficiency advantage is therefore reflected in both model size and computational cost. More importantly, the accuracy improvement is achieved without increasing model capacity or computation, showing that LBSNet does not rely on model scaling for better performance.

\subsection{Temporal Weighting Analysis}
\label{sec:t-SNE}
We examine how temporal weighting affects the aggregated features on SADT. In the shuffled setting, learned weights are randomly reassigned across temporal windows while their values remain unchanged. As shown in Fig.~\ref{fig:tsne}, alert and fatigue samples overlap substantially under shuffled weights, whereas learned weights yield clearer class separation. This difference shows that temporal weight allocation affects the aggregated features, consistent with the ablation results showing better classification with learned temporal aggregation. Together, these results support temporal weighting in forming more discriminative representations.

\subsection{Spectral Prototype Analysis}
\label{sec:Spectral_prototype_analysis}

We further analyze the learned spectral prototypes on SEED-VIG. As shown in Fig.~\ref{fig:prototype}(a), P1 has positive components in the theta, alpha, and beta bands and negative components in the delta and gamma bands, whereas P2 exhibits an approximately opposite pattern. These differences indicate that the prototypes capture distinct spectral configurations across frequency bands.
Fig.~\ref{fig:prototype}(b) relates these configurations to vigilance state, with P1 accounting for 65.6\% of the assignments in fatigue samples and P2 accounting for 71.6\% in alert samples. This association is also reflected in the continuous PERCLOS analysis in Fig.~\ref{fig:prototype}(c), where the proportion of P1 generally increases as PERCLOS rises, while that of P2 decreases.
Taken together, these results suggest that increasing fatigue is accompanied by a gradual redistribution between relative cross-band spectral states. This finding supports the use of prototype-guided temporal aggregation for fatigue modeling.

\section{Conclusion}
\label{sec:conclusion}

This work introduces LBSNet to connect local relative spectral patterns with temporal aggregation through learnable spectral prototypes, allowing spectral states to guide temporal weighting. Experiments on SEED-VIG and SADT show LBSNet outperforms all baselines while maintaining low computational cost. Learned prototypes reveal a gradual redistribution of relative cross-band spectral states with increasing fatigue, suggesting that joint spectral-state modeling offers a useful perspective for EEG-based fatigue assessment.

\vfill\pagebreak

\bibliographystyle{IEEEbib}
\bibliography{refs}

\end{document}